\documentclass{article}

\PassOptionsToPackage{numbers, compress}{natbib}

\usepackage[final]{nips_2018}

\usepackage[utf8]{inputenc} 
\usepackage[T1]{fontenc}    
\usepackage{hyperref}       
\usepackage{url}            
\usepackage{booktabs}       
\usepackage{amsfonts}       
\usepackage{nicefrac}       
\usepackage{microtype}      
\usepackage{graphicx}

\usepackage{amsmath,amssymb,amsfonts}
\usepackage{algorithmic}
\usepackage{textcomp}
\usepackage{xcolor}

\title{Multi-Task Generative Adversarial Network\\ for Handling Imbalanced Clinical Data}

\author{
  Mina Rezaei \\
  HPI, 
  Berlin, Germany \\
  \texttt{mina.rezaei@hpi.de} \\
  \And
  Haojin Yang \\
  HPI, 
  Berlin, Germany \\
  \texttt{haojin.yang@hpi.de} \\
  \And
  Christoph Meinel \\
  HPI, 
  Berlin, Germany \\
  \texttt{christoph.meinel@hpi.de} \\
}

\begin{document}

\maketitle

\begin{abstract}
We propose a new generative adversarial architecture to mitigate imbalance data problem for the task of medical image semantic segmentation where the majority of pixels belong to a healthy region and few belong to lesion or non-health region. A model trained with imbalanced data tends to bias towards healthy data which is not desired in clinical applications. We design a new conditional GAN with two components: a generative model and a discriminative model to mitigate imbalanced data problem through selective weighted loss. While the generator is trained on sequential magnetic resonance images (MRI) to learn semantic segmentation and disease classification, the discriminator classifies whether a generated output is real or fake. The proposed architecture achieved state-of-the-art results on ACDC-2017 for cardiac segmentation and diseases classification. We have achieved competitive results on BraTS-2017 for brain tumor segmentation and brain diseases classification.
\end{abstract}

\section{Introduction}

Medical imaging plays an important role in disease diagnosis, treatment planning, and clinical monitoring. One of the major challenges in medical image analysis is unbalanced data as normal or healthy data majority and lesion or non-healthy data are minor. A model learned from class imbalanced training data is biased towards the class with majority that is healthy. The predicted results of such networks have low sensitivity where sensitivity shows the ability of a test to correctly predict non-healthy classes. In medical applications, the cost of miss-classification of the minority class could be more than the cost of miss-classification of the majority class. For example, the risk of not detecting tumor could be much higher than referring to a healthy subject to doctors. 

The problem of class imbalanced have been recently addressed in diseases classification, tumor localization, and tumor segmentation and two types of approaches have been proposed in the literature: data-level approaches and algorithm-level approaches.

At the data-level, the objective is to balance the class distribution through re-sampling the data space~\cite{NIPS2000_1831} including SMOTE (Synthetic Minority Over-sampling Technique) of the positive class or under-sampling of the negative class~\cite{6914453}.
However, these approaches often lead to remove some important samples or add redundant samples to the training set.
Other techniques include patient-wise sampling~\cite{rezaei2018conditional} and incremental rectification of mini-batches for training deep neural network~\cite{dong2018imbalanced}.

Algorithm-level based solutions address class imbalanced problem by modifying the learning algorithm to alleviate the bias towards majority class. Examples are accuracy loss~\cite{sudre2017generalised}, Dice coefficient loss~\cite{rezaei2017deep}, and asymmetric similarity loss~\cite{D11078} that modify distribution of training data with regards to miss-classification cost. These losses are able to cover only some aspects of the quality of the application. For example in case of segmentation different measures such as mean surface distance or Hausdorff surface distance need to be used. In this work, we mitigate the negative impact of the class imbalance problem with new selective weighted loss to increase minority classes impact and decrease majority healthy class.

Moreover, we investigate the ability of proposed method for learning of semantic segmentation of body organ as well as abnormal tissues, with classification of diseases.
Our goal of multiple tasks learning is to utilize the correlation among these related tasks to improve the prediction of disease, besides the learning of the segmentation.
In our propose architecture, the generator network takes the 2D sequence multi-modal, multi-site medical images and outputs the 2D sequence of semantic segmentation of body organ (left ventricle, right ventricle, myocardium vessel) or abnormal tissues (brain tumor), and prediction of disease (brain disease or cardiac disease)
The discriminator takes the 2D sequence of generator output and 2D sequence of the ground truth annotated by an expert, to determine whether the generative output for segmentation is real or fake. The training procedure is a two-player mini-max game, where a generator network and a discriminator network are alternately trained to respectively minimize and maximize an objective function.
The adversarial loss is mixed with two additional losses: a weighted categorical cross-entropy loss to attenuate imbalanced data problem for semantic segmentation, and a weighted $\ell1$ loss for the task of diseases prediction.

\section{Method}
\begin{figure}[!t]
\includegraphics[width=0.86\textwidth]{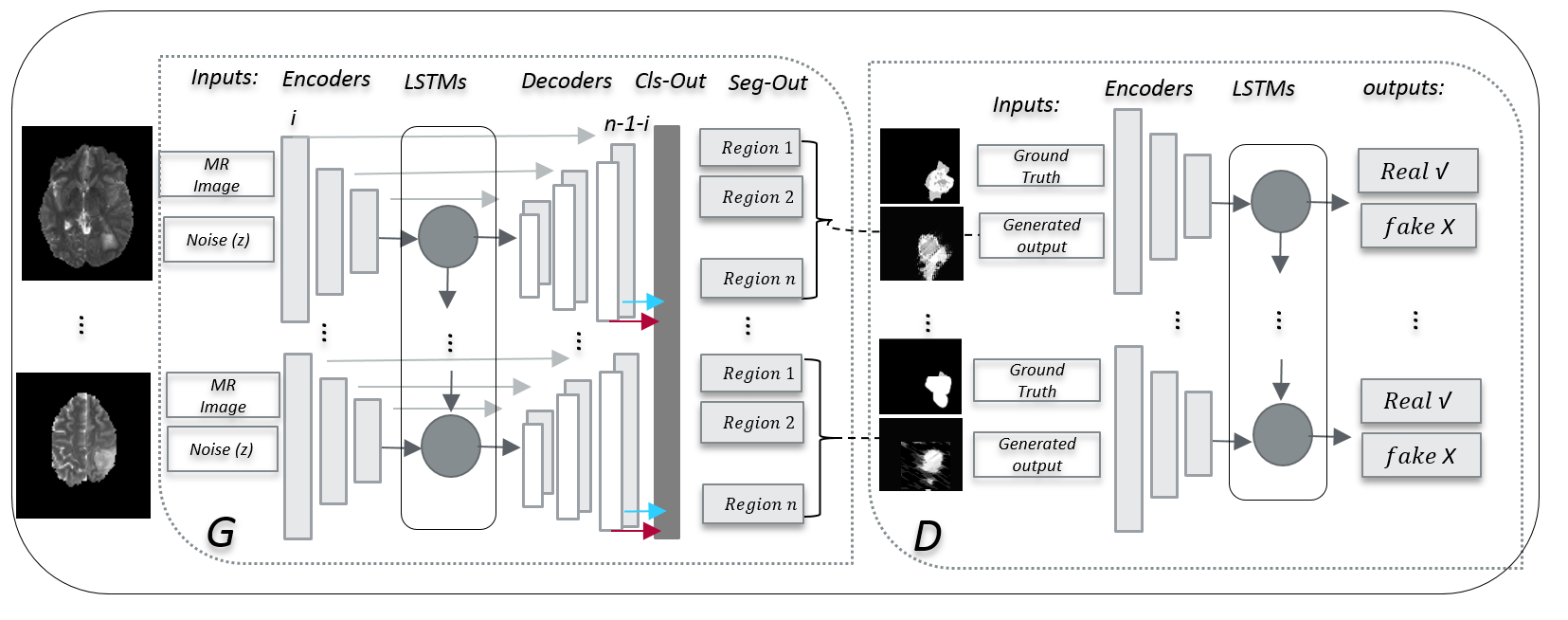}
\centering
\caption{Our proposed architecture for learning semantic segmentation and diseases prediction. We design a set of auto-encoders combined with a LSTM unit in a circumvent bottleneck as the generator network with skip connections between each layer i and the corresponding layer n-1-i (mostly like UNet architecture). The discriminator is fully convolutional network substituted with LSTM unit. Both networks are trained together in an adversarial way with selective weighted categorical cross entropy loss for semantic segmentation and selective weighted L1 for diseases prediction.}
\label{figarch}
\end{figure}

Similar conditional GAN~\cite{MirzaO14}; in our proposed method, a generative model learns mapping from a given sequence of 2D multimodal MR images $x_i$ to a sequence semantic segmentation $y_{i_seg}$ and classification $y_{cls}$; $G : \{x_i,z\} \rightarrow \{y_{i_seg}, y_{cls}\}$ (e.g. $i$ refers to 2D slice index between 1 and 155 from a total 155 slices acquired from each patient).
The training procedure for the segmentation task is similar to two-player mini-max game as shown in Eq.(\ref{eq_svGAN}).
While the generative model generated segmentation pixels label, the discriminator classifies whether the predicted pixel output by generator is similar the ground truth annotated by a medical expert or synthetic.
\begin{equation} \label{eq_svGAN}
\mathcal{L}_{adv} \leftarrow \underset{G} min ,  \underset{D } max , V(G, D) = E_{x,y_{seg}} [log D(x,y_{seg})] + E_{x,z} [log (1-D(x, G(x,z)))]
\end{equation}
A class-imbalanced in a medical dataset where non-healthy classes could not be trained as well as healthy classes, might dominate the gradient direction.
Regarding to mitigate class-imbalanced impact, we mixed adversarial loss~\ref{eq_svGAN} with selective weighted~(Eq.\ref{SW}) categorical cross-entropy loss~$\mathcal{L}_{H}(G)$ for semantic segmentation, and with selective weighted~(Eq.\ref{SW}) $\ell1$ loss~Eq.(\ref{L1}) for classification of diseases.
\begin{equation} \label{SW}
w_c = \sqrt{ \dfrac{ \mid C_c \mid }{f_c + N} }
\end{equation}
Where the weight for each class $c$ is based on the ratio of the cordiality among $N$ classes on entire training dataset (e.g. mostly healthy classes) by the frequency of samples with class $c$ appears in the dataset. Since we have intense frequency differences, the square root is applied to prevent huge weights. This implies that larger classes in the training set have a weight smaller than 1 and the weights of the smallest classes are the highest defined by Eq.(\ref{SW}). The final loss for semantic segmentation task is calculated through Eq.(\ref{seg})
\begin{equation} \label{seg}
\mathcal{L}_{seg} (D, G) = \mathcal{L}_{adv} (D, G) +  \mathcal{L}_{H}(G * w_c) 
\end{equation}
As shown in Fig.\ref{figarch}, the concatenated depth features from last decoder layer with skip connection of encoder part passed into a couple of dense layers and map to the disease class. The objective function for class prediction is $\ell1$ to minimize the absolute difference between the predicted value and the existing largest value Eq.(\ref{L1}) 
\begin{equation} \label{L1}
 \mathcal{L}_{cls}(G) = E_{x} \parallel y_{cls} -  \sum\limits_{i=1} G(x_i * w_c) \parallel 
\end{equation}
Where $i$ indicates to 2D slice index from the same patient (e.g. in Brain dataset $i$ is between 1 and 155 from a total of 155 slices acquired from each patient).
The final objective function for simultaneous semantic segmentation and classification is:
\begin{equation} \label{Lfinal}
 \mathcal{L} = \mathcal{L}_{seg} (D, G) +  \mathcal{L}_{cls}(G)
\end{equation}

\section{Experiments} \label{gen_inst}
We tested the performance of our proposed method for simultaneous segmentation and classification on two different clinical datasets: BraTS-2017~\cite{Menze2014,Bakasnature2017,Bakastcg2017,Bakaslgg2017} in two different brain diseases and four semantic segmentation labels belong to different tumor regions. The ACDC-2017~\cite{bernard2018deep} dataset consists of five subgroups cardiac diseases and four segmentation labels belong to cardiac regions.

\textbf{ Data Preparation and Augmentation:} The pre-processing is an important step in medical image analysis to bring all subjects in similar distributions. We applied z-score normalization on the both datasets. We also applied bias field correction, as standard scale normalization introduced by Ny{\'u}l et al.~\cite{nyul2000new}.
To prevent over fitting, the patient-wise batch normalization~\cite{rezaei2017conditional}, and data augmentation such as randomly cropped, re-sizing, scaling, rotation between -10 and 10 degree, and Gaussian noise applied on training and testing time for both datasets.

\textbf{ Configuration:} Our proposed method is implemented based on a Keras library~\cite{chollet2015keras} with back-end Tensorflow~\cite{abadi2016tensorflow} and is publicly available on authors github~\footnote{https://github.com/HPI-DeepLearning/SV-GAN}. The learning rate is initially set to 0.0001. The RMSprop optimizer is used in both the generator and the discriminator, it dividing the learning rate by an exponentially decaying average of squared gradients. The model is trained for up to 120 epochs.

\textbf{ Network Architecture:} In this work, a generative network is a modified UNet architecture consist of ten fully convolutional layers and four max-pooling layers where a bidirectional LSTM unit is fed in the circumvent bottleneck.
The UNet architecture allows low-level features to shortcut across the network. The bidirectional LSTM provides inter as intra slice feature representation which is very important in sequential medical image analysis. The advantage of bidirectional LSTM appear when we connected features from $n-1-i$ and $i$ (as shown in Fig~\ref{figarch}) from same patient and passed them into fully connected layer to classify patient diseases. Our discriminator is fully convolutional Markovian PatchGAN classifier~\cite{Phillipimagetoimage2017} which only penalizes structure at the scale of image patches.
Unlike, the PathGAN discriminator introduced by Isola et al.~\cite{Phillipimagetoimage2017} which classified each N $\times$ N patch for real or fake, we have achieved better results for task of semantic segmentation in pixel level where we consider N=1. Moreover, since we have a sequential data, the bidirectional LSTM added after last CNN layer in discriminator network. Same as encoder part of generator architecture, the discriminator consists of five fully convolutional layers and four max-pooling layers. 

\begin{figure}
\includegraphics[width=0.50\textwidth]{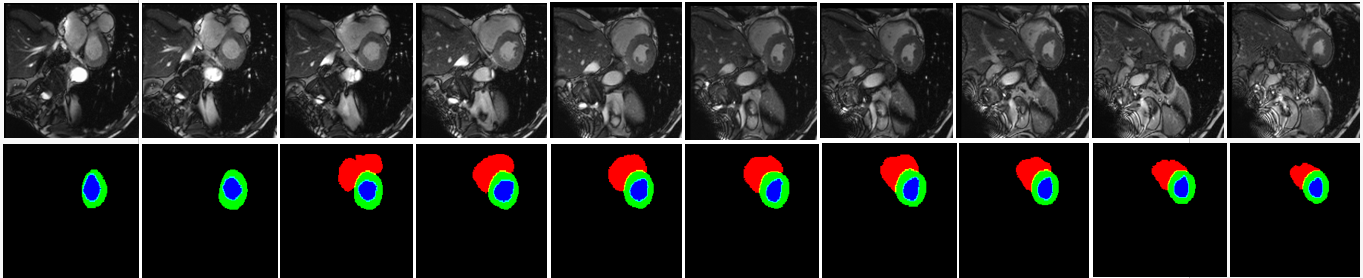}
\centering
\caption{Segmentation results obtained by our proposed method on ACDC-2017. First row shows input images and second row shows the semantic segmentation output where the red, blue and green colors respectively are the right, left ventricles, and myocardium vessel.}
\label{fig_validation}
\end{figure}

We use binary cross entropy as an adversarial loss, weighted categorical cross entropy as an additional loss for generative model for task of semantic segmentation. The $\ell1$ loss is for classification.
Regarding the heavy class imbalance in both datasets, minority classes might not be trained as well as majority classes specially in the task of semantic segmentation, which we used selective weighted categorical cross entropy loss as segmentation loss. From Table~\ref{tableSegResults}, second row provide results without weighted loss, while first row in both Tables show the obtained results with selective weighted loss. Fig.~\ref{fig_validation} shows qualitative results on ACDC-2017 benchmark.
In this work, the recurrent architecture selected for both discriminator and generator is a bidirectional LSTM proposed by Graves et al.~\cite{graves2005framewise}.

\textbf{Evaluation}
We evaluated the performance of our approach on 3D MRI for semantic segmentation using the quality metrics introduced by the ACDC-2017 challenge organizer~\cite{bernard2018deep} and online judgment system of BraTS-2017. Similar to the winners of BraTS 2017~\cite{kamnitsas2017ensembles,wang2017automatic}, we do tumorous segmentation by attenuating imbalanced data effect, where they trained models in cascade architectures but we applied weighted loss function.
Compared to~\cite{kamnitsas2017ensembles,wang2017automatic}, our proposed network has the advantage of carrying out multiple clinical tasks in a single architecture where we have achieved 98.61\% accuracy for diagnosis between high and low grade glioma tumor. It is important to mention that our method takes only 58 seconds to segment one MR brain image consisting 155 slices at testing time.

\begin{table} [http]
\caption{The first part of this Table shows our achieved accuracy for semantic segmentation in terms of Dice, Hausdorff distance (Hdff), and Sensitivity (Sen) on unseen BraTS-2017 data and comparison with related and top rank approaches. The WT, ET, and TC columns respectively are abbreviation of whole tumorous region, enhanced tumorous, and tumorous active core. The second part of Table is comparison and achieved accuracy in term of Dice metric and Hausdorff distance in average of end of systolic and end of diastolic phase from ACDC benchmark with related approaches and top ranked methods reported in~\cite{bernard2018deep}. The LV, RV, and MYO columns respectively are abbreviation of left ventricle region, right ventricle region, and myocardium vessel. the second column shows the accuracy results in diseases classification.}
\begin{center}
\begin{tabular}{|c|c|c|c|c|c|c|c|c|c|c|}
\hline
\text{}&\text{Cls}&\multicolumn{3}{|c|}{\text{Dice}} &\multicolumn{3}{|c|}{\text{Hausdorff}} &\multicolumn{3}{|c|}{\text{Sensitivity}}\\
\cline{3-11} 
\text{Architecture}&\text{Acc} & \text{\textit{WT}}& \text{\textit{ET}}& \text{\textit{TC}} & \text{\textit{WT}}& \text{\textit{ET}}& \text{\textit{TC}} & \text{\textit{WT}}& \text{\textit{ET}}& \text{\textit{TC}}\\
\hline
Weighted-RNN-GAN & 98 & 0.88 & 0.76 & 0.77 & 6.11 & 7.17 & 11.38  & 0.87 & 0.88 & 0.85  \\
pix-to-pix & - & 0.80 & 0.61  & 0.61 & 7.30  & 9.22 & 12.04 & 0.75 & 0.61 &  0.55  \\
SegAN~\cite{XueXZLH17} & -. & 0.85 & 0.66 & 0.70 & - & - &  - & 0.80 & 0.62 & 0.65  \\
3D UNet+ FCN~\cite{kamnitsas2017ensembles} & - & 0.90 & 0.74 & 0.79 & 4.23 & 4.5 & 6.56 & 0.89 & 0.78 & 0.86 \\
Cascade 2D UNet~\cite{wang2017automatic} & - & 0.90 & 0.78 & 0.83 & 3.89 & 3.28 &  6.48 & 0.91 & 0.77 & 0.82 \\
\hline
\text{}&\text{Cls}&\multicolumn{3}{|c|}{\text{Dice}} &\multicolumn{3}{|c|}{\text{Hausdorff}} &\multicolumn{3}{|c|}{\text{Sensitivity}}\\
\cline{3-11} 
\text{Architecture}&\text{Acc} & \text{\textit{LV}}& \text{\textit{RV}}& \text{\textit{MYO}} & \text{\textit{LV}}& \text{\textit{RV}}& \text{\textit{MYO}} & \text{\textit{LV}}& \text{\textit{RV}}& \text{\textit{MYO}}\\
\hline
Weighted-RNN-GAN & 94 & 0.95 & 0.93 & 0.92 & 8.11 & 8.17 & 8.38  & - & - & -  \\
pix-to-pix & - & 0.9 & 0.88 & 0.87 & 9.34 & 13.1 & 9.18  & - & - & -  \\
Dilated CNN~\cite{wolterink2017automatic} & 84 & 0.95 & 0.93 & 0.89 & 8.8 & 12.2 &  10.5 &  - & - & - \\
\hline
\end{tabular}
\label{tableSegResults}
\end{center}
\end{table}

\section{Conclusion}
\label{headings}
In this paper, we introduced a new adversarial framework for learning simultaneous semantic segmentation and diseases prediction. We developed and evaluated the selective weighted loss to mitigate the issue of imbalanced data for medical image analysis.
To this end, we proposed a generator network and a discriminator network where a generator generates pixel's label, and discriminator classifies whether generator output is real or fake. We achieved promising results on two popular medical imaging benchmarks for the task of semantic segmentation of abnormal tissues as well as a body organ, together with a prediction of diseases.
In the future, we plan to investigate the potential of multiple agents GANs for learning of multiple clinical tasks.

\small
\bibliographystyle{nips.bst}
\bibliography{refs.bib}

\end{document}